\documentclass[letterpaper, 10 pt, conference]{ieeeconf}  % Comment this line out if you need a4paper

\IEEEoverridecommandlockouts                              % This command is only needed if 
\usepackage{amsmath}
\usepackage{graphicx}      % include this line if your document contains figures
\usepackage{svg}
\usepackage{amsmath, amssymb}
\usepackage{graphicx}
\usepackage{caption}
\usepackage{amsmath, amssymb}
\usepackage{graphicx}
\usepackage{svg}
\usepackage{caption}

\usepackage{amsmath}

\usepackage{pifont}
\newcommand{\cmark}{\ding{51}} % tick
\newcommand{\xmark}{\ding{55}} % cross

\usepackage{float}                 % VERY IMPORTANT
\usepackage[ruled,vlined]{algorithm2e}
\usepackage{cite}
\usepackage{listings}
\usepackage{xcolor}
\title{\LARGE \bf
RoboNav-Arm: Agentic AI-Driven Navigation and Obstacle Avoidance for Robotic Manipulator in Cluttered Environments
}

\author{Aachal Sharma$^{1}$ and Narendra Kumar Dhar$^{1}$%
\thanks{$^{1}$Centre for Artificial Intelligence and Robotics (CAIR), Indian Institute of Technology Mandi, Himachal Pradesh 175005, India.
        {\tt\small}}%
}
\usepackage{tcolorbox}
\tcbuselibrary{listings,skins,breakable}
\usepackage{graphicx}
\usepackage{subcaption}
\usepackage{xcolor}
\usepackage{caption}

\definecolor{codebg}{RGB}{245,245,245}
\definecolor{keywordcolor}{RGB}{0,0,180}
\definecolor{stringcolor}{RGB}{163,21,21}

\tcbset{
  colback=codebg,
  colframe=black,
  listing only,
  breakable,
  enhanced
}

\begin{document}

\maketitle

%%%%%%%%%%%%%%%%%%%%%%%%%%%%%%%%%%%%%%%%%%%%%%%%%%%%%%%%%%%%%%%%%%%%%%%%%%%%%%%%

\begin{abstract}
Robotic manipulators operating in unstructured environments face significant challenges in safely executing goal-directed tasks due to dynamic and unforeseen obstacles, while traditional methods rely on prior knowledge or fixed perception pipelines, limiting adaptability. 
We propose a framework for safe task execution with effective obstacle avoidance.
The environment module performs real-time obstacle detection, 3D localization, and ground surface geometry estimation. It then generates a structured semantic report that includes obstacle positions, object geometry and shape, and whether obstacles lie inside, outside, or within critical interaction zones.
A central coordination module manages the overall system by handling tool invocation (e.g., memory and MoveIt collision scene updates), facilitating communication between modules, and continuously monitoring task progress until completion.
Furthermore, a planning module selects an appropriate motion planning algorithm, such as RRTConnect, RRT*, or BiTRRT, based on the current environment configuration and goal requirements. The trajectory generated by the planner is further analyzed and refined to ensure safe and collision-free task execution.
The proposed approach is evaluated in Gazebo Classic , demonstrating robustness in dynamic scenarios.
\end{abstract}

%%%%%%%%%%%%%%%%%%%%%%%%%%%%%%%%%%%%%%%%%%%%%%%%%%%%%%%%%%%%%%%%%%%%%%%%%%%%%%%%
\section{Introduction }

Robotic manipulation has witnessed significant advancements in recent years, allowing robots to deal with more complex tasks with better accuracy and efficiency. However, working effectively in cluttered, obstacle-rich environment remains a difficult problem. In such situations, robots have to adjust their movements carefully, especially when performing tasks like pick-and-place, where they must move around obstacles while dealing with changing and uncertain conditions.
Obstacle avoidance is not only about preventing collisions, but also about keeping a safe distance from obstacles, since even getting too close can affect both safety and reliability. At the same time, changes in the environment and uncertainties during execution make things more challenging, so robots need to keep adjusting their actions as they operate. For this reason, obstacle avoidance is better viewed as a dynamic and context-dependent problem rather than a fixed planning task.
 \begin{figure}[!t]
    \centering
    \includegraphics[
        width=\columnwidth,
        height=0.7\textheight,
        keepaspectratio
    ]{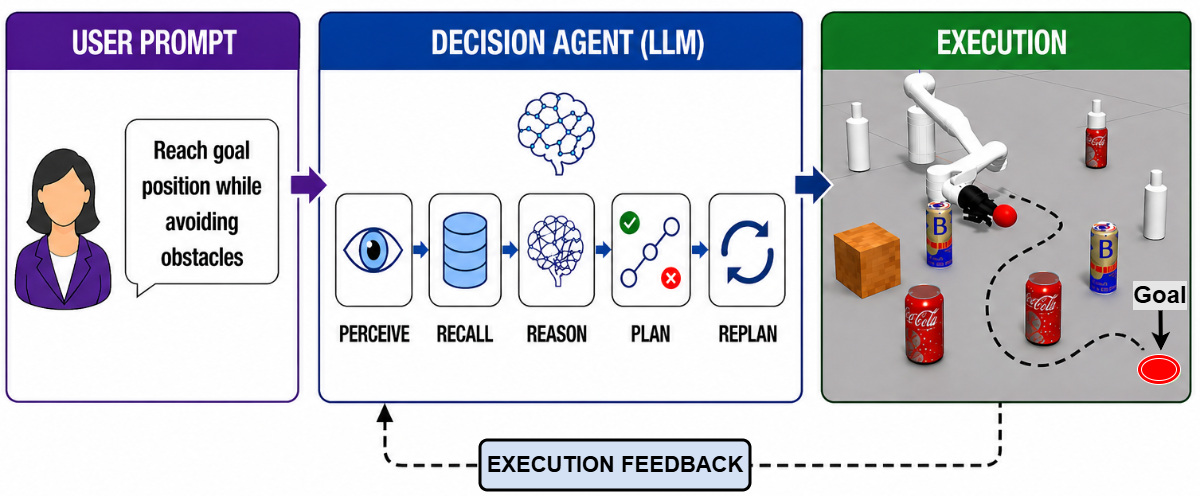}
    \caption{An overview of RoboNav-Arm}
\label{fig:robonav_arm_overview}
\end{figure}
Classical motion planning methods, including sampling-based approaches for instance, methods such as Rapidly-exploring Random Trees (RRT) and Probabilistic Roadmaps (PRM) along with their improved versions and task-motion planning frameworks, have been widely used for obstacle avoidance in robotic manipulation \cite{shen2023manipulability, islam2024trajectory,zhang2025rrtstar}. However, they assume full prior environmental knowledge and predefined planning strategies, limiting their use in unknown environments.

To deal with these limitations, learning-based approaches such as reinforcement learning (RL) and imitation learning have been studied. RL-based methods allow robots to learn through interaction, which is useful for complex manipulation tasks, but they often struggle with low sample efficiency and the challenge of defining suitable reward functions \cite{meng2025rlpathplanning}, \cite{jiang2025world4rl}. On the other hand, imitation learning makes use of expert demonstrations and has performed well in more structured manipulation settings \cite{wang2025imitationframework}. Even so, these approaches typically require large amounts of training data, involve significant computational cost, and may not always be reliable in safety-critical scenarios. In addition, they do not explicitly account for safety margins or the risks associated with operating in close proximity to obstacles.

Over the past few years, large language models (LLMs) have seen rapid progress, and this has led to growing interest in using agentic AI for robotic systems. These approaches can support reasoning, break tasks into smaller steps, and assist in high-level decision-making, which allows robots to generate action sequences from natural language instructions \cite{song2023llmplanner, cheng2024affordanceprompting, chen2025vlmpc}. Alongside this, work in embodied AI and vision–language–action (VLA) models indicates that perception, reasoning, and control can be combined more closely within a single framework \cite{kannan2024smartllm}. Methods such as VoxPoser, as well as related approaches, show that language models can also be used to guide trajectory generation by linking semantic understanding with spatial representations \cite{huang2023voxposer}, \cite{kwon2024trajectorylm}.

Even with recent progress, most LLM-based systems still focus mainly on high-level planning and depend on external motion planners or predefined primitives when it comes to execution. For this reason, they do not really deal with the low-level motion planning issues involved in obstacle avoidance. Another issue is that LLMs can generate outputs that are not always feasible or safe, particularly in long-horizon tasks, and they often assume correct execution without accounting for uncertainties in the environment or possible failures. They also lack explicit mechanisms for safety-aware reasoning, including handling near-collision scenarios and real-time adaptation.

As a result, no existing approach is able to simultaneously provide (i) high-level reasoning, (ii) low-level motion planning, and (iii) safety-aware real-time adaptation in unknown environments within a single framework.

In this work, we propose an agentic AI-based framework for obstacle-aware robotic manipulation as illustrated in Fig.~\ref{fig:robonav_arm_overview}. The system is built around a decision-making agent and uses a set of modular components to handle environment understanding, planner selection, and trajectory execution. A safety-aware validation step is included to check not only for collisions but also for cases where the robot operates too close to obstacles. The framework also makes use of continuous monitoring during execution, so that feedback can be used to adjust actions and re-plan when required.

By integrating reasoning, safety-aware motion planning, and continuous adaptation within a unified framework, the proposed approach enables robust and reliable obstacle avoidance in dynamic robotic manipulation scenarios.
The key contribution of our work is
\begin{itemize}
\item We propose an end-to-end agentic AI framework that integrates perception and planning for obstacle avoidance in robotic manipulation tasks.
\item We develop an environment perception agent capable of estimating 3D object positions, ground surface geometry, and per-object workspace classification, enabling direct integration with the MoveIt collision scene.
\item We evaluate the proposed framework across diverse scenarios with varying obstacle densities, demonstrating its effectiveness and robustness in unknown environments.
\end{itemize}
Table~\ref{tab:key_comparison} presents a comparison of the proposed method with state-of-the-art approaches.

The rest of the paper is organized as follows: Section II describes the proposed methodology. Section III presents the experimental setup, results, and discussion. Section IV concludes the paper.

\begin{table}[htbp]
\centering
\renewcommand{\arraystretch}{1.2}
\small

\resizebox{\columnwidth}{!}{%
\begin{tabular}{|l|c|c|c|c|c|}
\hline
\textbf{Feature} & \textbf{\cite{11093514}} & \textbf{\cite{wang2024llm}} & \textbf{\cite{11043104}} & \textbf{\cite{11163014}} & \textbf{Proposed} \\
\hline
Agentic Reasoning & \cmark & \cmark & \xmark & \xmark & \cmark \\
\hline
Object-Level Scene Understanding & \cmark & \xmark &\xmark & $\triangle$ & \cmark \\
\hline
Memory-Guided Adaptation & \xmark & $\triangle$ & \xmark & \xmark & \cmark \\
\hline
Collision-Aware Validation & \xmark & $\triangle$  & $\triangle$  & $\triangle$  & \cmark \\
\hline
\end{tabular}
}

\caption{Comparison of the proposed method with state-of-the-art robotic manipulation and obstacle avoidance approaches ($\triangle$ indicates partial support).}
\label{tab:key_comparison}
\end{table}

\begin{figure*}[h]
\centering
\includegraphics[width=\linewidth]{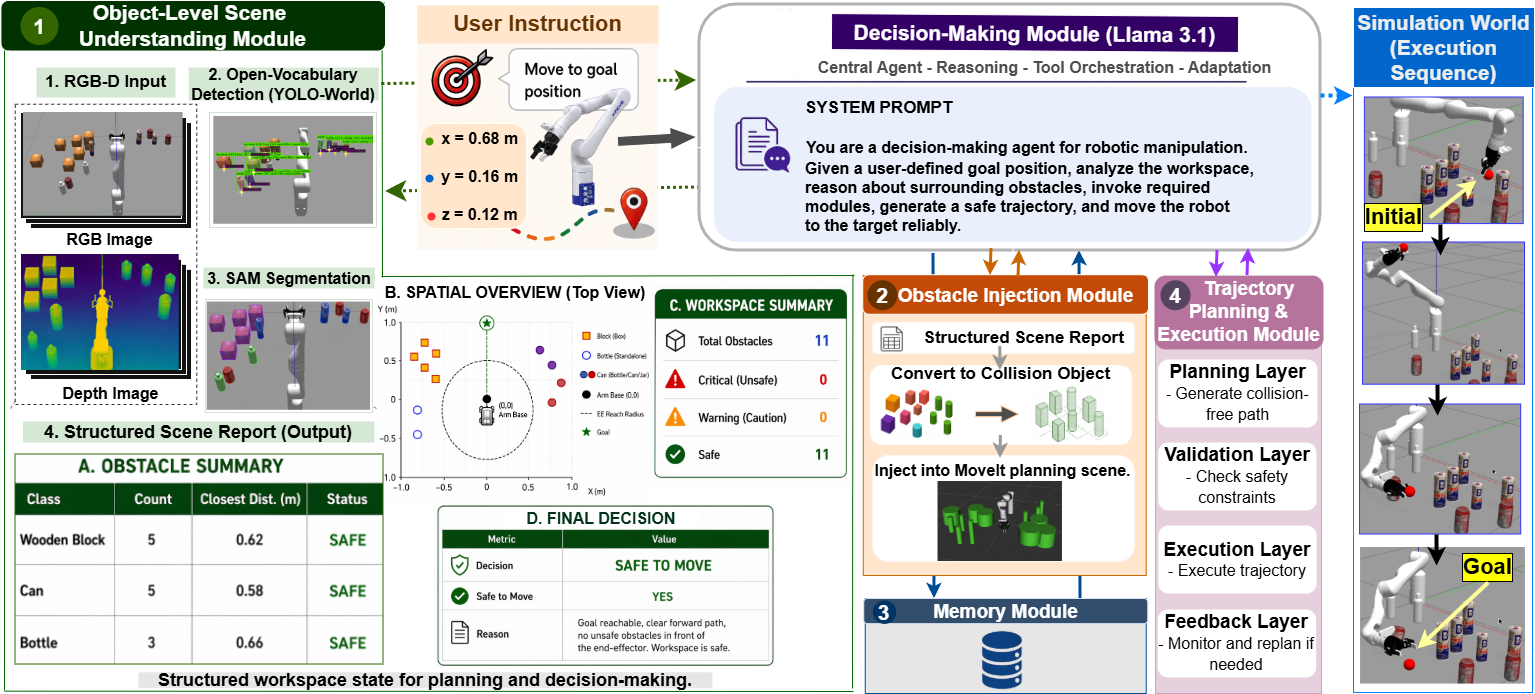}
\caption{The overall architecture of RoboNav-Arm, which consists of different tools used by the decision-making agent.}
\label{fig:robonav_arm_architecture}
\end{figure*}

\section{Methodology}
\subsection{Agentic Framework for Collision-Aware Planning}

We propose an agentic framework in which a decision-making agent performs high-level reasoning, adaptive module selection, and execution orchestration for robotic manipulation with integrated obstacle avoidance, as shown in Fig.~\ref{fig:robonav_arm_architecture}. The primary challenge
lies in adaptively coordinating perception, contextual memory, and motion planning under dynamic and geometrically
constrained workspace conditions.

The framework is designed to address the following key challenges:

\begin{itemize}

\item \textit{How can the system construct collision-aware planning representations from complex workspace observations?}

\item \textit{How can prior execution experience improve planning robustness and adaptive decision-making under uncertain conditions?}

\item \textit{How can safe and reliable trajectory execution be maintained through continuous monitoring and replanning?}

\end{itemize}

To address these challenges, the proposed framework is formally defined as:
\[
\mathcal{D}
=
\langle
\mathcal{G},
\mathcal{S},
\mathcal{A}
\rangle
\]
where $\mathcal{G}$ denotes the goal position, $\mathcal{S}=\{\theta_t,\mathcal{E}_t\}$ represents the workspace state consisting of the current manipulator joint configuration $\theta_t$ and the environment representation $\mathcal{E}_t=\{\mathcal{O}_t,\mathcal{W}_t\}$ returned by the Object-Level Scene Understanding Module, and $\mathcal{A}$ corresponds to the set of high-level operational modules available to the decision-making agent.

The available high-level operational modules are represented as
\[
\mathcal{A}
=
\{
a_{env},
a_{mem},
a_{plan}
\}
\]
where $a_{env}$ invokes the Object-Level Scene Understanding Module (OSUM), $a_{plan}$ initializes the Trajectory Planning and Execution Module (TPEM), and $a_{mem}$ queries the Contextual Memory and Retrieval Module (CMRM). 

At each decision step, the decision-making agent generates a high-level operational decision formulated as
\[
d_t
=
\mathrm{LLM}
\left(
a_t,
s_t,
h_t
\right),
\quad
a_t \in \mathcal{A},
\;
s_t \in \mathcal{S}
\]
where $d_t$ denotes the action generated by the reasoning agent, $s_t$ represents the current workspace state, and $h_t$ denotes the accumulated execution history and contextual reasoning information.

The proposed framework utilizes a large language model (LLM) as a high-level reasoning agent responsible for module orchestration and adaptive planning decisions using the  available operational modules, current workspace state, and accumulated execution context. 

The agent initially receives the desired goal pose and invokes OSUM to obtain a planning-aware representation of the environment. Upon receiving $\mathcal{E}_t$, the detected object geometries are inserted as collision objects into the MoveIt planning scene, thereby updating the planning environment to accurately reflect current spatial conditions for collision-aware trajectory generation.

The decision-making agent then evaluates environmental complexity and determines whether additional contextual reasoning through CMRM is required before initiating planning. Under execution-critical conditions such as obstacle-dense environments, repeated planning failures, reduced obstacle clearance, or geometrically constrained configurations, the agent selectively retrieves relevant execution context and previously successful planning strategies associated with similar environments.

Using the combined environment representation and retrieved contextual information, the agent determines planner selection, safety-aware execution parameters, and obstacle avoidance strategies before invoking TPEM. The planning module subsequently generates and executes a collision-aware manipulator trajectory while the decision-making agent maintains supervisory control over execution behavior.

% During execution, the framework continuously monitors trajectory safety and the current workspace state. If unsafe conditions, planning failures, or unexpected workspace changes are detected, the decision-making agent re-evaluates the decision phase by invoking the OSUM to update workspace understanding, querying the CMRM for relevant prior experience, and adapting planning behavior through the TPEM to trigger replanning under updated workspace conditions.

% In this manner, the proposed framework establishes a closed-loop agentic manipulation system in which the reasoning agent continuously observes environmental conditions, evaluates execution feedback, selectively invokes computational modules, and adaptively refines planning behavior to achieve robust obstacle-aware robotic manipulation.
During execution, the framework continuously monitors trajectory safety and the current workspace state. If unsafe conditions, planning failures, or unexpected workspace changes are detected, the decision-making agent re-evaluates the decision phase by invoking the OSUM for updated scene understanding, querying the CMRM for relevant prior experience, and triggering replanning through the TPEM.

In this manner, the proposed framework establishes a closed-loop agentic manipulation system for obstacle-aware robotic manipulation in which the decision-making agent continuously observes the workspace state, evaluates execution feedback, selectively invokes operational modules, and adaptively refines planning decisions for robust obstacle-aware robotic manipulation.

\begin{tcolorbox}[
title=\textbf{Decision-Making Prompt},
colframe=green!40!black,
colback=green!2,
width=\columnwidth,
sharp corners,
boxrule=0.1pt]

\begin{lstlisting}[
basicstyle=\ttfamily\scriptsize,
aboveskip=1pt,
belowskip=-3pt,
columns=fullflexible,
breaklines=true,
showstringspaces=false]
ROLE: Motion planning agent
INPUT: obstacle_count, min_obstacle_distance, goal_pose, previous_failures
TASK: analyze workspace constraints, select planner, set velocity, adapt planning strategy
OUTPUT: {planner:"...", velocity:..., safety_strategy:"...", reasoning:"..."}
\end{lstlisting}

\end{tcolorbox}
\subsubsection{Object-Level Scene Understanding Module}
The decision-making agent requires structured and physically grounded understanding of the environment to support safe and adaptive manipulation. In obstacle-aware execution, reasoning depends on accurate knowledge of object geometry, spatial layout, and workspace constraints.

To address this, an \textbf{Object-Level Scene Understanding Module} is introduced as a perception component that transforms raw sensory observations into a structured, planning-compatible workspace representation.

The module utilizes RGB-D observations to generate an object-centered representation of the workspace. Instead of directly relying on raw perception outputs, each object is represented through its spatial and geometric attributes to support downstream reasoning and motion planning.

Objects are initially detected using YOLO-World~\cite{cheng2024yoloworld}, enabling open-vocabulary recognition of objects treated as workspace obstacles without restricting perception to predefined categories. The detected object category defines the semantic object label \( l_i \), while the resulting bounding boxes provide coarse object localization.

The detected object regions are further refined using instance-level segmentation through SAM~\cite{kirillov2023sam} to obtain pixel-accurate object masks. This segmentation process removes background regions and preserves only object-specific observations for subsequent geometric analysis.

The segmented object regions are projected into 3D using depth observations to generate reconstructed object point clouds \( P_i \) represented in the robot base coordinate frame.

The environment representation is formulated as
\[
\mathcal{E}_t
=
\{
\mathcal{O}_t,
\mathcal{W}_t
\}
\]
% where \( \mathcal{O}_t \) denotes the object-level scene representation and \( \mathcal{W}_t \) represents the workspace-level reasoning context utilized for planning-aware decision-making.
% The object-level representation is defined as
% \[
% \mathcal{O}_t
% =
% \{
% o_1,o_2,\dots,o_n
% \}
% \]

% where each object is represented as
% \[
% o_i
% =
% (l_i,d_i,g_i,p_i)
% \]
% with \( l_i \) denoting the semantic object label obtained from object detection, \( d_i \) representing object dimensions estimated by computing an orientation-aware bounding volume aligned with the principal geometric axes of the reconstructed point cloud, \( g_i \) denoting the geometric properties inferred as
where \( \mathcal{O}_t = \{o_1,o_2,\dots,o_n\} \) denotes the object-level scene representation and \( \mathcal{W}_t \) represents the workspace-level reasoning context utilized for planning-aware decision-making. Each object is represented as \( o_i = (l_i,\Delta_i,g_i,p_i) \), where \( l_i \) denotes the semantic object label, \( \Delta_i \) represents object dimensions estimated from the reconstructed point cloud geometry, \( g_i \) denotes geometric properties inferred as
\[
g_i
=
\mathrm{LLM}(l_i,\Delta_i)
\]
where the language model utilizes \( l_i \) and \( \Delta_i \) to generate compact collision representations such as boxes and cylinders, while \( p_i \) denotes the estimated object position.
For a reconstructed object point cloud \( P_i \), the ground-support surface region corresponding to the physical contact between the object and the supporting plane is estimated as
\[
B_i
=
\{
p \in P_i
\mid
z(p)
\leq
z_{\min}
+
\epsilon
\}
\]
where \( z_{\min} \) denotes the minimum vertical coordinate of the reconstructed object points and \( \epsilon \) is a small tolerance used to account for surface irregularities.

The object position \( p_i \) is then estimated from the centroid of the extracted ground-support surface region:
\[
p_i
=
\frac{1}{|B_i|}
\sum_{p \in B_i} p
\]
This estimation provides a physically stable and geometrically accurate object position by utilizing the lowest supporting surface region instead of centroid-based localization.

The generated object representations are aggregated into the structured scene representation \( \mathcal{E}_t \), where the workspace-level reasoning context \( \mathcal{W}_t \) is generated through LLM-based reasoning over object geometry, obstacle distribution, free-space structure, and workspace constraints. The returned scene representation provides planning-relevant information required for obstacle analysis, geometric reasoning, adaptive planning, and collision-aware manipulation under dynamically changing workspace conditions.

\begin{tcolorbox}[
title=\textbf{Scene Analysis Prompt},
colframe=blue!70!black,
colback=blue!3,
width=\columnwidth,
sharp corners,
boxrule=0.1pt]

\begin{lstlisting}[
basicstyle=\ttfamily\scriptsize,
aboveskip=1pt,
belowskip=-3pt,
columns=fullflexible,
breaklines=true,
showstringspaces=false]
ROLE: Environment analysis agent
INPUT: object positions, size/shape, end-effector distance
TASK: analyze obstacles, classify risk, check workspace safety
OUTPUT: {obstacle_analysis:[...], workspace_summary:"...", safe_to_move:True/False}
\end{lstlisting}

\end{tcolorbox}
\subsubsection{Trajectory Planning and Execution Module}

% The decision-making agent needs a way to turn high-level goals and structured scene information into actual robot movements. In obstacle-aware manipulation, generating a path alone is not enough. The system must also make sure the movement is feasible, safe, and suitable for the current environment. 

% To address this, a \textbf{Trajectory Planning and Execution Module} is introduced as an execution tool operating under the control of the decision-making agent.

% The module takes as input the goal pose, current robot state, and planning configuration provided by the decision-making agent. Since the planning scene, including collision objects and their geometric properties, has already been constructed externally, it directly interfaces with the motion planner to generate a candidate trajectory.

Once the planning-compatible workspace representation is generated, the decision-making agent invokes the \textbf{Trajectory Planning and Execution Module} for motion generation and safety-aware trajectory validation. Based on the selected planning strategy and workspace constraints, the planner generates a manipulator trajectory represented as
\[
\tau
=
\{
Q_1,Q_2,\dots,Q_n
\}
\]
where each trajectory point \( Q_k = (\theta_1,\theta_2,\dots,\theta_n) \) corresponds to a valid \( n \)-DOF manipulator joint configuration generated during motion planning.

To improve execution safety under constrained workspace conditions, obstacle geometries are expanded using safety-aware inflation:
\[
 g_i'
=
g_i 
\oplus
B(\sigma_i)
\]
where \( g_i \) denotes the  geometry of the perceived object \( O_i \), \( B(\sigma_i) \) represents a safety buffer of radius \( d_{safe} \), and \( \oplus \) denotes the Minkowski sum operation. The safety margin is determined as
\[
\sigma_i
=
f(g_i,\lambda_i)
\]
where  \( \lambda_i \) corresponds to additional execution clearance introduced for robust obstacle avoidance. The resulting inflated collision representation is integrated into the MoveIt planning scene for collision checking and trajectory validation. For every trajectory point \( Q_k \in \tau \), the corresponding manipulator geometry is represented as \( R(Q_k) \), which defines the occupied robot volume at configuration \( Q_k \).
Collision-free trajectory validation is then performed using
\[
\forall Q_k \in \tau,
\quad
R(Q_k)\cap g_i' = \emptyset
\]
ensuring that every manipulator configuration along the generated trajectory remains collision-free with respect to the safety-expanded obstacle representation. Only trajectories satisfying the required safety constraints are approved for execution through the robot control interface, enabling safe and physically consistent robot motion under constrained workspace conditions.

During execution, the decision-making agent continuously supervises trajectory safety and obstacle proximity using the minimum obstacle clearance
\[
\phi
=
\min_i
\|R(Q_k)-g_i'\|
\]
where  \( \|\cdot\| \) computes the minimum distance between the manipulator geometry and surrounding collision objects. If the minimum obstacle clearance violates the predefined safety threshold \( \phi < \delta \), the current trajectory execution is terminated, and replanning is dynamically triggered under updated workspace conditions.

\subsubsection{Contextual Memory and Retrieval Module}
To avoid repeatedly reasoning over geometrically related workspace conditions, we propose a \textbf{Contextual Memory and Retrieval Module} in our framework that stores prior manipulation experience from previous task executions. The stored memory representation is formulated as
\[
\mathcal{M}
=
\{
(
\mathcal{E}_i,
\tau_{p_i},
c_i,
r_i
)
\}_{i=1}^{N}
\]
where \( \tau_{p_{i}} \) denotes the successful previous trajectories, \( c_{i} \) denotes the execution-safety context information, \( r_{i}\) denotes the recovery and replanning strategies from previous execution, and \( N \) denotes the total number of stored memory entries.

During execution, relevant prior experiences are retrieved via vector similarity search using ChromaDB.
\[
m_r
=
\arg\max_{m_i \in \mathcal{M}}
\kappa(s_t,m_i)
\]
where ( $\kappa(\cdot) $) measures geometric similarity between the current workspace configuration and stored memory entries. For workspace conditions that are close to previously encountered scenarios, the system retrieves relevant past experience and uses it to support planner selection and replanning with strategies that worked in earlier attempts.

% ???????????????????????????????????????????????????????????????????????????????????????

% \begin{tcolorbox}[
%     title=\textbf{Structured Scene Report},
%     colframe=blue!70!black,
%     colback=blue!5,
%     width=\columnwidth,
%     sharp corners,
%     boxrule=0.8pt
% ]

% \begin{lstlisting}[
%     basicstyle=\ttfamily\scriptsize,
%     columns=fullflexible,
%     breaklines=true,
%     showstringspaces=false
% ]
% SCENE:
%   obstacle_density: 0.72 (high)
%   min_clearance: 0.034 m
%   workspace_bounds: [-0.5, 0.5, -0.5, 0.5, 0.0, 0.8] m
%   robot_pose: [0.0, 0.0, 0.0, 0.0, 0.0, 0.0, 1.0]

% OBJECTS:
%   [0] cup
%       shape: cylinder
%       position: [0.32, -0.15, 0.00] m (bottom surface)
%       dimensions: [r=0.04, h=0.10] m

%   [1] book
%       shape: box
%       position: [0.45, 0.10, 0.00] m
%       dimensions: [l=0.20, w=0.15, h=0.02] m

% ROBOT:
%   current_pose: [0.0, 0.0, 0.5, 0.0, 0.0, 0.0]
%   reachable: True
% \end{lstlisting}

% \end{tcolorbox}

% ///////////////////////////////////////////////////////////////////////////////////

\section{EXPERIMENTAL SETUP AND EVALUATION SCENARIOS}
\begin{figure}[htbp]
    \centering
    \includegraphics[
        width=\columnwidth,
        height=0.55\textheight,
        keepaspectratio
    ]{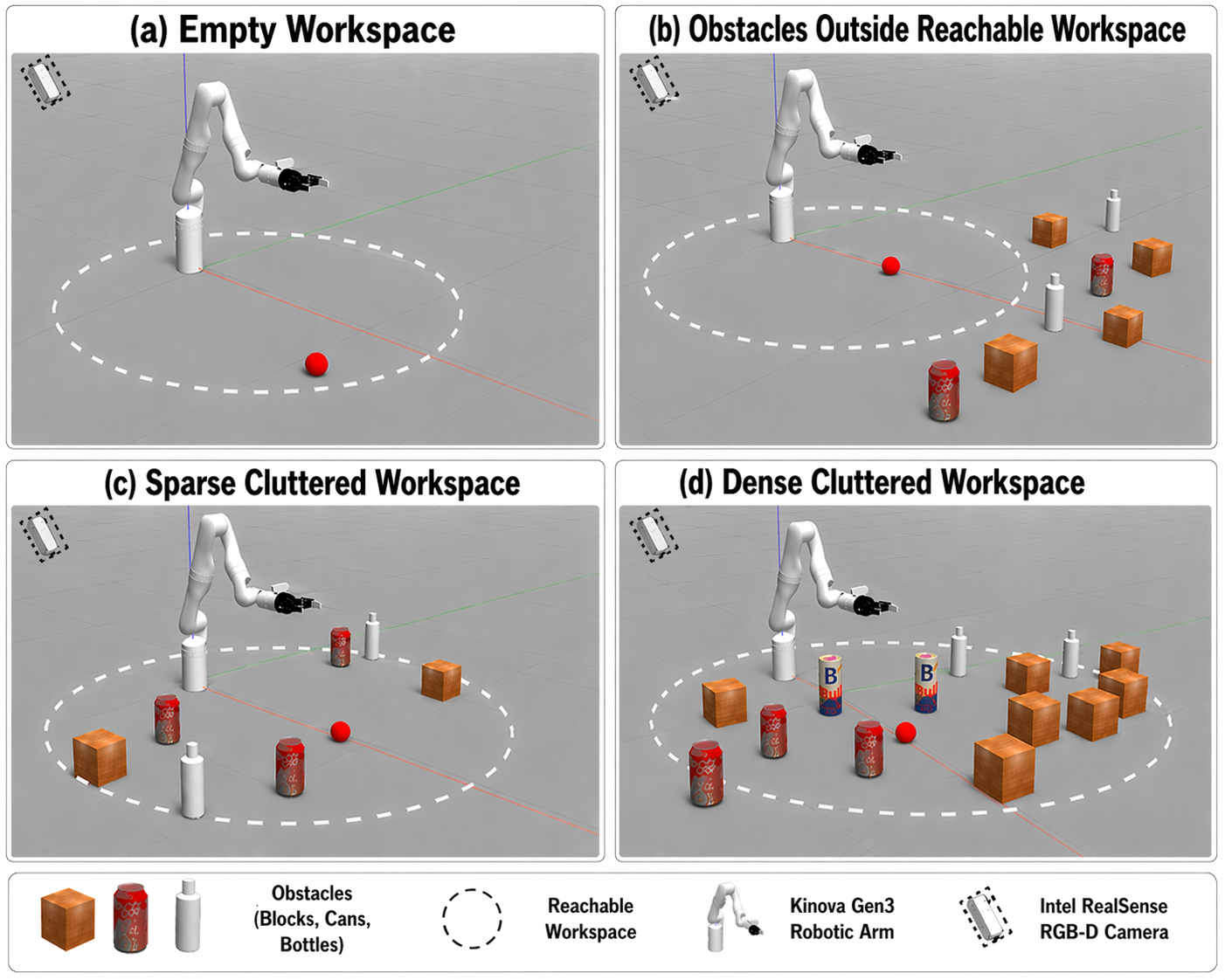}
    \caption{Experimental setup and evaluation scenarios}
    \label{fig:EnvSetup}
\end{figure}
All experiments were conducted using a Kinova Gen3 7-DOF robotic manipulator equipped with a Robotiq 2F-85 parallel-jaw gripper. An RGB-D camera was deployed in a fixed eye-to-hand configuration, mounted externally in the world frame to provide full workspace visibility.  The calibrated camera pose in the world frame was $(-0.55, 0.0, 0.85)$ m with orientation $(0, 0.698, 0)$ rad. Four workspace configurations with increasing spatial complexity were designed, as illustrated in Fig. \ref{fig:EnvSetup}.

\textbf{Scenario 1 (S1): Empty Workspace} No obstacles were present inside the workspace, allowing free robot motion, as shown in Fig.~\ref{fig:EnvSetup}(a).
\textbf{Scenario 2 (S2): Obstacles Outside Reachable Workspace} Obstacles were placed outside the robot's reachable workspace and did not affect motion, as shown in Fig.~\ref{fig:EnvSetup}(b).
\textbf{Scenario 3 (S3): Sparse Cluttered Workspace} Some obstacles were outside the workspace, while most were inside the reachable region with sufficient spacing, creating a sparse cluttered environment, as shown in Fig.~\ref{fig:EnvSetup}(c).
\textbf{Scenario 4 (S4): Dense Cluttered Workspace} Most obstacles were placed inside the workspace and close to each other, while a few remained outside, creating a dense and challenging environment, as shown in Fig.~\ref{fig:EnvSetup}(d).

\section{Result and discussion}
\begin{figure}[htbp]
    \centering
    \includegraphics[
        width=\columnwidth,
        height=0.35\textheight,
        keepaspectratio
    ]{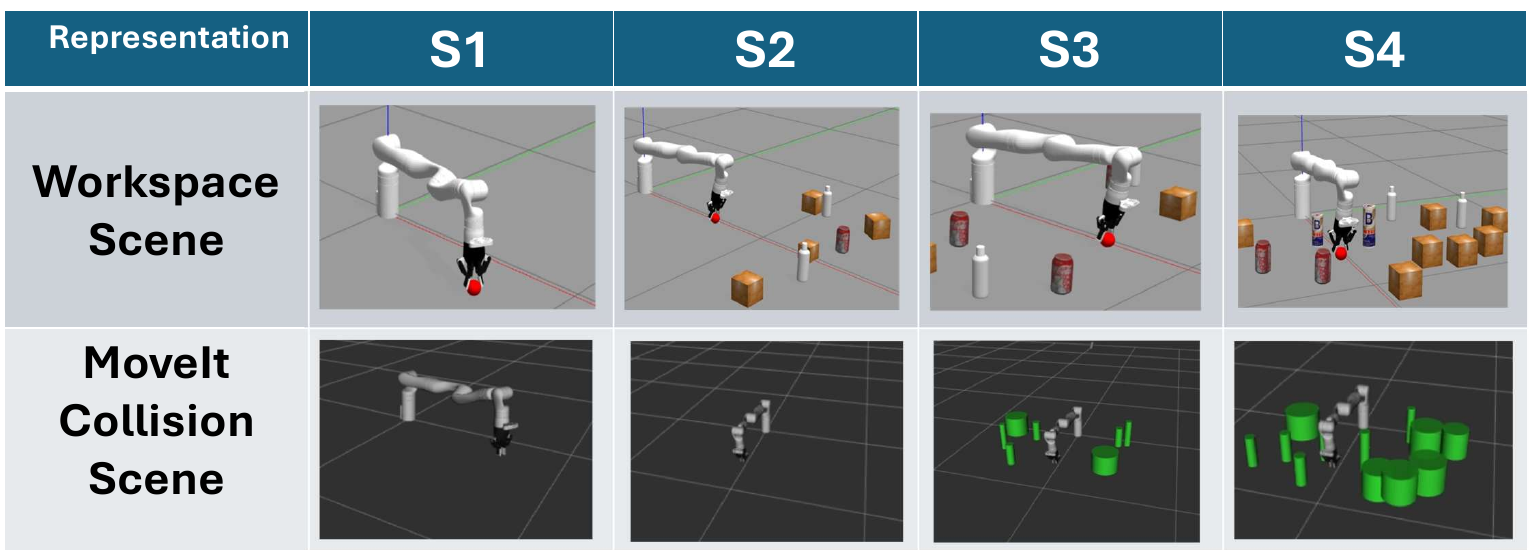}
    \caption{Visualization of workspace configurations and corresponding MoveIt collision environments under different obstacle arrangements, showing consistent mapping between the actual workspace and the injected planning environment.}
    \label{fig:CollisionScene}
\end{figure}

\begin{table}[t]
\centering

\renewcommand{\arraystretch}{1.0}
\setlength{\tabcolsep}{5.5pt}

\begin{tabular}{|c|c|c|c|}
\hline
\textbf{Metric} & \textbf{DeepSeek (S3/S4)} & \textbf{Qwen (S3/S4)} & \textbf{Llama (S3/S4)} \\
\hline
\shortstack{Planning\\Time (s)}
& \shortstack{61.0$\pm$31.1 /\\89.0$\pm$39.1}
& \shortstack{62.9$\pm$29.7 /\\105.0$\pm$18.4}
& \shortstack{{37.5$\pm$26.0 /}\\59.6$\pm$30.4} \\
\hline

\shortstack{Execution\\Time (s)}
& \shortstack{108.0$\pm$45.8 /\\170.69$\pm$77.46}
& \shortstack{110.1$\pm$40.5 /\\182.35$\pm$57.65}
& \shortstack{73.3$\pm$18.7 /\\93.1$\pm$31.0} \\
\hline

\shortstack{Trajectory\\Points}
& \shortstack{53.9$\pm$15.2 /\\59.8$\pm$16.13}
& \shortstack{55.4$\pm$10.0 /\\68.2$\pm$18.96}
& \shortstack{58.7$\pm$17.8 /\\68.5$\pm$15.7} \\
\hline
\shortstack{ No of \\
Replans}
& \shortstack{2.5$\pm$1.1 /\\3.6$\pm$1.26}
& \shortstack{2.6$\pm$1.0 /\\3.9$\pm$0.74}
& \shortstack{2.0$\pm$1.1 /\\2.9$\pm$1.3} \\
\hline
\end{tabular}
\caption{Comparative performance evaluation of different LLM  for our method S3 and S4 workspace configurations.}
\label{tab:llm_comparison}
\end{table}
All experiments were conducted under the same setup, where the robot moved to a predefined fixed initial position before each trial and the grasping configuration was kept constant for all scenarios. The desired goal position (x, y, z) was entered by the user through the GUI for each test.
Fig.~\ref{fig:CollisionScene} illustrates the actual workspace scenes along with their corresponding MoveIt collision scenes generated through the OSUM. This demonstrates that the detected obstacles from the real environment were accurately mapped and injected into the planning scene for collision-aware motion planning under each obstacle configuration.

To analyze the performance of the proposed framework under different LLMs, comparative experiments were conducted using DeepSeek-V4-Flash (DeepSeek), Qwen-480B-A35B-Instruct (Qwen), and Llama-3.1-70B-Instruct (Llama) under two obstacle-rich scenarios (S3 and S4). Table~\ref{tab:llm_comparison} summarizes the mean and standard deviation values of planning time, execution time, trajectory complexity, and replanning behavior across repeated trials. Among the evaluated LLMs, Llama demonstrated comparatively lower planning and execution overhead with more stable replanning behavior across both workspace conditions. Based on the comparative evaluation, Llama was selected for the detailed experiments and overall framework validation presented in the following sections.

\begin{figure}[htbp]
\centering
% Row 1
\begin{subfigure}{0.20\textwidth}
    \centering
    \includegraphics[width=\linewidth]{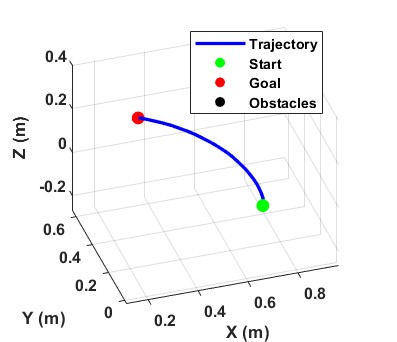}
    \caption{}
\end{subfigure}
\hspace{0.005\textwidth}
\begin{subfigure}{0.22\textwidth}
    \centering
    \includegraphics[width=\linewidth]{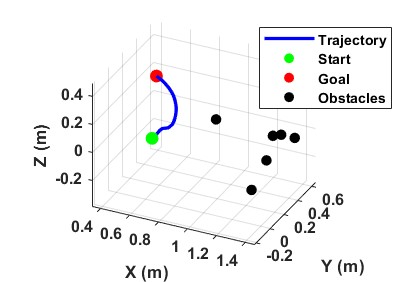}
    \caption{}
\end{subfigure}
\vspace{-0.4em}
\begin{subfigure}{0.22\textwidth}
    \centering
    \includegraphics[width=\linewidth]{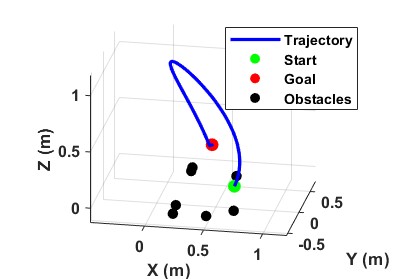}
    \caption{}
\end{subfigure}
\hspace{0.005\textwidth}
\begin{subfigure}{0.20\textwidth}
    \centering
    \includegraphics[width=\linewidth]{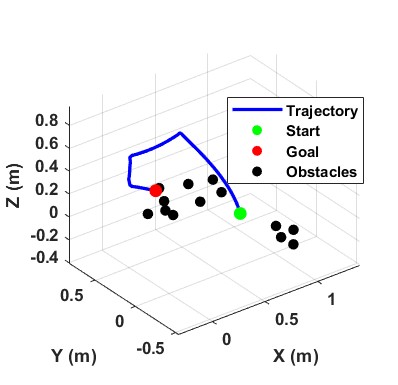}
    \caption{}
\end{subfigure}
\caption{Robot trajectory with obstacle configurations for four experimental scenarios: (a) S1, (b) S2, (c) S3, and (d) S4.}
\label{fig:trajectory_results_4}
\end{figure}

\begin{table}[h]
\centering
\renewcommand{\arraystretch}{1.05}
\small
\resizebox{\columnwidth}{!}{%
\begin{tabular}{|c|c|c|c|c|c|}
\hline
\shortstack{\textbf{Test}\\\textbf{Scenario}} & 
\shortstack{\textbf{Planning}\\\textbf{Time (s)}} & 
\shortstack{\textbf{Execution}\\\textbf{Time (s)}} & 
\shortstack{\textbf{Trajectory}\\\textbf{Points}} & 
\shortstack{\textbf{No. of}\\\textbf{Replans}} \\
\hline

\shortstack{Scenario 1} & $18.9 \pm 9.6$ & $45.8 \pm 11.6$  &  $42.4 \pm 11.5$ &  $1.1 \pm 0.3$ \\
\hline
\shortstack{Scenario 2} & $24.7 \pm 10.4~$ & $46.1 \pm 13.8~$  & $48.3 \pm 14.3$ & $1.2 \pm 0.4$ \\
\hline
\shortstack{Scenario 3} & $37.5 \pm 26.0$ & $78.3 \pm 18.7$ & $58.7 \pm 17.8$ & $2.0 \pm 1.1$ \\
\hline
\shortstack{Scenario 4} & $59.6 \pm 30.4$ & $93.1 \pm 31.0$ & $68.5 \pm 15.7$ & $2.9 \pm 1.3$ \\
\hline

\end{tabular}
}
\caption{Performance Evaluation Across Different Workspace Scenarios}
\label{tab:scenario_results}
\end{table}
Table~\ref{tab:scenario_results} presents the mean values with standard deviations of planning time, execution time, path length, and the number of times a new planner was selected by the agent during multiple trials for each workspace scenario. In the Empty Environment case, lower planning effort, shorter trajectories, and minimal agent intervention were observed due to the absence of obstacles and collision constraints. This simple setup also achieved a perfect collision-free success rate of 100\%, as shown in Table~\ref{tab:success_rate}. In the Obstacles Outside Workspace scenario, where some obstacles were positioned close to but outside the workspace boundary, only a slight increase in planning effort, path length, and replanning frequency was observed, while the success rate remained unchanged at 100\%. As workspace complexity increased in the sparse and dense cluttered scenarios, longer paths, higher execution times, and more frequent planner changes were required to generate safe trajectories. Consequently, the collision-free success rate decreased to 93.3\% and 83.3\%, respectively, as summarized in Table~\ref{tab:success_rate}, due to the increased navigation difficulty in constrained environments.
\begin{table}[h]
\centering
\renewcommand{\arraystretch}{0.85}
% \footnotesize
\small
\resizebox{\columnwidth}{!}{%
\begin{tabular}{|c|c|c|c|}
\hline
\shortstack{\textbf{Test}\\\textbf{Scenario}} & 
\shortstack{\textbf{Trials}} & 
\shortstack{\textbf{Successful}\\\textbf{Runs}} & 
\shortstack{\textbf{Collision-Free}\\\textbf{Success Rate (\%)}} \\
\hline
\shortstack{Scenario 1} & 30 & 30 & 100 \\
\hline

\shortstack{Scenario 2} & 30 & 30 & 100 \\
\hline

\shortstack{Scenario 3} & 30 & 28 & 93.3 \\
\hline

\shortstack{Scenario 4} & 30 & 25 & 83.3 \\
\hline

\end{tabular}
}
\caption{Collision-free success rate of RoboNav-Arm under different workspace scenarios.}
\label{tab:success_rate}
\end{table}
These experimental results were obtained from multiple trials using different goal positions within the reachable workspace. For visualization purposes, Fig. \ref{fig:trajectory_results_4} presents one representative goal position at $(0.36,\ 0.64,\ 0.10)$ to show the robot trajectory from the predefined initial position across all four scenarios.

By validation, we say that overall, the proposed approach demonstrated robust and reliable performance during experimental testing, successfully adapting to different workspace complexities while maintaining collision-free task execution.

% These results demonstrate the ability of the proposed framework to adapt its planning strategy and maintain reliable collision-free task execution under increasing workspace complexity.

% Table~\ref{tab:scenario_results} presents the average planning time, execution time, path length, and number of times a new planner is chosen by the agent reaching a collision-free goal during multiple trials for each workspace scenario. In the empty environment and obstacle-free relevant workspace cases, lower planning effort, shorter trajectories, and minimal agent intervention were observed due to fewer motion constraints. As obstacle density increased in the sparse and dense cluttered scenarios, longer paths, higher execution time, additional replanning, and more frequent agent interventions were required to maintain collision-free motion. These results demonstrate the ability of the proposed framework to adapt its planning strategy and execute tasks reliably under increasing workspace complexity.

\section{Conclusion}

We presented RoboNav-Arm, an end-to-end agentic framework for obstacle avoidance in robotic manipulators operating in cluttered and unknown environments. The proposed system combines real-time scene understanding, memory-assisted reasoning, adaptive planner selection, and safety-aware trajectory execution to achieve reliable and collision-free motion planning. Validation in Gazebo Classic across multiple workspace scenarios demonstrates strong robustness under varying obstacle densities and workspace complexities. The proposed framework enhances scalability, adaptability, and seamless integration of additional capabilities, making it well-suited for next-generation intelligent robotic systems. Future work will focus on handling dynamic moving obstacles, improving real-time re-planning, and incorporating learning-driven decision-making for more autonomous operation.

% Row 1

\bibliographystyle{IEEEtran}
\bibliography{reference}

@article{shen2023manipulability,
  author    = {Henghua Shen and Wen-Fang Xie and Jianyu Tang and Tao Zhou},
  title     = {Adaptive Manipulability-Based Path Planning Strategy for Industrial Robot Manipulators},
  journal   = {IEEE/ASME Transactions on Mechatronics},
  volume    = {28},
  number    = {3},
  pages     = {1742--1753},
  year      = {2023},
  doi       = {10.1109/TMECH.2022.3231467}
}

@article{islam2024trajectory,
  author    = {Naeem Ul Islam and Kaynat Gul and Faizullah Faiz and Syed Sajid Ullah and Ikram Syed},
  title     = {Trajectory Optimization and Obstacle Avoidance of Autonomous Robot Using Robust and Efficient Rapidly Exploring Random Tree},
  journal   = {PLoS ONE},
  volume    = {19},
  number    = {10},
  pages     = {e0311179},
  year      = {2024},
  doi       = {10.1371/journal.pone.0311179}
}

@article{zhang2025rrtstar,
  author    = {Xiqing Zhang and Pengyu Wang and Yongrui Guo and Qianqian Han and Kuoran Zhang},
  title     = {Path Planning Algorithm for Manipulators in Complex Scenes Based on Improved RRT*},
  journal   = {Sensors},
  volume    = {25},
  number    = {2},
  pages     = {328},
  year      = {2025},
  doi       = {10.3390/s25020328}
}

@inproceedings{meng2025rlpathplanning,
  author    = {Zhijian Meng and Xin Zhou and Hui Chen and Jianliang Mao},
  title     = {Reinforcement Learning-based Robotic Arm Path Planning in Complex Scenarios},
  booktitle = {Proceedings of the 2nd International Conference on Computer, Internet of Things and Smart City (CIoTSC)},
  pages     = {167--172},
  year      = {2025},
  publisher = {ACM},
  doi       = {10.1145/3731867.3731895}
}

@article{jiang2025world4rl,
  author    = {Zhennan Jiang and Kai Liu and Yuxin Qin and Shuai Tian and Yupeng Zheng and Mingcai Zhou and Chao Yu and Haoran Li and Dongbin Zhao},
  title     = {World4RL: Diffusion World Models for Policy Refinement with Reinforcement Learning for Robotic Manipulation},
  journal   = {arXiv preprint arXiv:2509.19080},
  year      = {2025},
  doi       = {10.48550/arXiv.2509.19080}
}

@article{wang2025imitationframework,
  author    = {Weiyong Wang and Chao Zeng and Hong Zhan and Chenguang Yang},
  title     = {A Novel Robust Imitation Learning Framework for Complex Skills With Limited Demonstrations},
  journal   = {IEEE Transactions on Automation Science and Engineering},
  volume    = {22},
  pages     = {3947--3959},
  year      = {2025},
  doi       = {10.1109/TASE.2024.3403833}
}

@inproceedings{song2023llmplanner,
  author    = {Chan Song and B. M. Sadler and Jiaman Wu and Wei-Lun Chao and Clayton Washington and Yu Su},
  title     = {LLM-Planner: Few-Shot Grounded Planning for Embodied Agents with Large Language Models},
  booktitle = {Proceedings of the IEEE International Conference on Computer Vision (ICCV)},
  pages     = {2986--2997},
  year      = {2023},
  doi       = {10.1109/ICCV51070.2023.00280}
}

@inproceedings{kannan2024smartllm,
  author    = {Shyam Kannan and Vishnunandan Venkatesh and Byung-Cheol Min},
  title     = {SMART-LLM: Smart Multi-Agent Robot Task Planning using Large Language Models},
  booktitle = {Proceedings of the IEEE/RSJ International Conference on Intelligent Robots and Systems (IROS)},
  pages     = {12140--12147},
  year      = {2024},
  doi       = {10.1109/IROS58592.2024.10802322}
}

@article{cheng2024affordanceprompting,
  author    = {Guangran Cheng and Chuheng Zhang and Wenzhe Cai and Li Zhao and Changyin Sun and Jiang Bian},
  title     = {Empowering Large Language Models on Robotic Manipulation with Affordance Prompting},
  journal   = {arXiv preprint arXiv:2404.11027},
  year      = {2024}
}

@article{chen2025vlmpc,
  author    = {Jiaming Chen and Wentao Zhao and Ziyu Meng and Donghui Mao and Ran Song and Wei Pan and Wei Zhang},
  title     = {Vision-Language Model Predictive Control for Manipulation Planning and Trajectory Generation},
  journal   = {arXiv preprint arXiv:2504.05225},
  year      = {2025},
  url       = {https://arxiv.org/abs/2504.05225}
}

@article{huang2023voxposer,
  author    = {Wenlong Huang and Chen Wang and Ruohan Zhang and Yunzhu Li and Jiajun Wu and Li Fei-Fei},
  title     = {VoxPoser: Composable 3D Value Maps for Robotic Manipulation with Language Models},
  journal   = {arXiv preprint arXiv:2307.05973},
  year      = {2023},
  doi       = {10.48550/arXiv.2307.05973}
}

@article{kwon2024trajectorylm,
  author    = {Teyun Kwon and Norman Di Palo and Edward Johns},
  title     = {Language Models as Zero-Shot Trajectory Generators},
  journal   = {IEEE Robotics and Automation Letters},
  volume    = {9},
  number    = {7},
  pages     = {6728--6735},
  year      = {2024},
  doi       = {10.1109/LRA.2024.3410155}
}

@inproceedings{
wang2024llm,
title={{LLM}3: Large Language Model-based Task and Motion Planning with Motion Failure Reasoning},
author={Shu Wang and Muzhi Han and Ziyuan Jiao and Zeyu Zhang and Ying Nian Wu and Song-Chun Zhu and Hangxin Liu},
booktitle={Multi-modal Foundation Model meets Embodied AI Workshop @ ICML2024},
year={2024},
url={https://openreview.net/forum?id=55UkYJLTs8}
}

@INPROCEEDINGS{11093514,
  author={Guo, Sheng and Tian, Tan and Zhou, JinXian and Liu, Zhiyang and Zhou, Lei and Zhao, Ruiteng and Huang, Zefan and Yuan, Chengran and Ang, Marcelo H},
  booktitle={2025 10th International Conference on Control and Robotics Engineering (ICCRE)}, 
  title={An End-to-End GPT-4o Informed Pipeline for Manipulation Task Planning with Sim2Real Validation}, 
  year={2025},
  volume={},
  number={},
  pages={99-104},
  doi={10.1109/ICCRE65455.2025.11093514}}

@INPROCEEDINGS{11163014,
  author={Porée, R. and Mujica, M. and Tricot, N. and Cadenat, V.},
  booktitle={2025 European Conference on Mobile Robots (ECMR)}, 
  title={Non-Holonomic Mobile Manipulator NMPC for Occlusion Avoidance Based on Elliptic Cone FOV Representation}, 
  year={2025},
  volume={},
  number={},
  pages={1-7},
  doi={10.1109/ECMR65884.2025.11163014}}

@INPROCEEDINGS{11043104,
  author={Vinaud Neto, Lázaro Pereira and Rodrigues, Nicolas Carreiro and Santos, Bernardo Rodrigues Tameirão and Do Valle Simões, Eduardo},
  booktitle={2025 IEEE Congress on Evolutionary Computation (CEC)}, 
  title={Evolutionary Computation Applied to the Control of a Robotic Manipulator with Obstacle Avoidance}, 
  year={2025},
  volume={},
  number={},
  pages={1-4},
  doi={10.1109/CEC65147.2025.11043104}}

@inproceedings{cheng2024yoloworld,
  author    = {Cheng, Tianheng and Song, Lin and Ge, Yixiao and Liu, Wenyu and Wang, Xinggang},
  title     = {YOLO-World: Real-Time Open-Vocabulary Object Detection},
  booktitle = {Proceedings of the IEEE/CVF Conference on Computer Vision and Pattern Recognition (CVPR)},
  pages     = {16901--16911},
  year      = {2024},
  doi       = {10.1109/CVPR52733.2024.01599}
}

@article{kirillov2023sam,
  author  = {Kirillov, Alexander and Mintun, Eric and Ravi, Nikhila and Mao, Hanzi and Rolland, Chloe and Gustafson, Laura and Xiao, Tete and Whitehead, Spencer and Berg, Alexander C. and Lo, Wan-Yen and Doll{\'a}r, Piotr and Girshick, Ross},
  title   = {Segment Anything},
  journal = {arXiv preprint arXiv:2304.02643},
  year    = {2023},
  doi     = {10.48550/arXiv.2304.02643}
}
% \section*{ACKNOWLEDGMENT}

% The preferred spelling of the word ÒacknowledgmentÓ in America is without an ÒeÓ after the ÒgÓ. Avoid the stilted expression, ÒOne of us (R. B. G.) thanks . . .Ó  Instead, try ÒR. B. G. thanksÓ. Put sponsor acknowledgments in the unnumbered footnote on the first page.
\end{document}